\documentclass[10pt,twocolumn,letterpaper]{article}

\usepackage{cvpr}
\usepackage{times}
\usepackage{epsfig}
\usepackage{graphicx}
\usepackage{amsmath}
\usepackage{amssymb}
\usepackage{comment}
\usepackage{caption}
\usepackage{stfloats}

\usepackage{booktabs}
\usepackage{array}
\newcolumntype{P}[1]{>{\centering\arraybackslash}p{#1}}
\newcolumntype{M}[1]{>{\centering\arraybackslash}m{#1}}
\usepackage[pagebackref=true,breaklinks=true,letterpaper=true,colorlinks,bookmarks=false]{hyperref}

\cvprfinalcopy 

\def\cvprPaperID{xxxx} 

\begin{document}

\newcommand{\OURS}{RevealNet}

\title{\OURS: Seeing Behind Objects in RGB-D Scans}

\author{
\parbox{4cm}{\centering Ji Hou\quad}
\parbox{4cm}{\centering Angela Dai\quad}
\parbox{4cm}{\centering Matthias Nie{\ss}ner}\\[0.3em]
Technical University of Munich
\vspace{0.4cm}
}

\twocolumn[{%
\renewcommand\twocolumn[1][]{#1}%
\maketitle

\begin{center}
\includegraphics[width=1.0\linewidth]{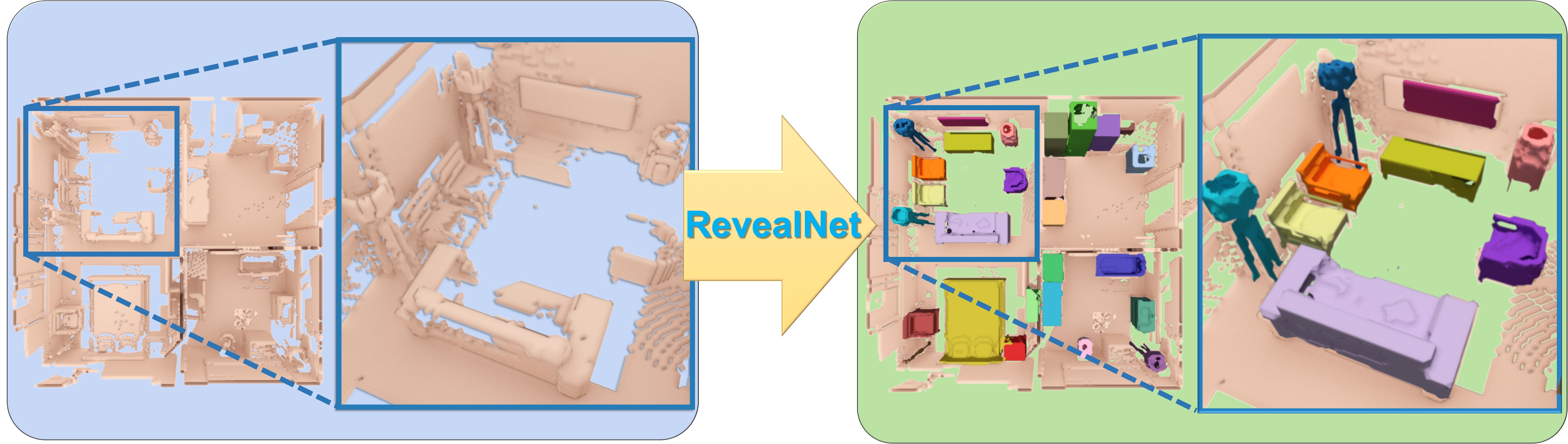}
\captionof{figure}{\OURS{} takes an RGB-D scan as input and learns to ``see behind objects'': from the scan's color images and geometry (encoded as a TSDF), objects in the observed scene are detected (as 3D bounding boxes and class labels) and for each object, the complete geometry of that object is predicted as per-instance masks (in both seen and unseen regions).
}
\vspace{0.4cm}
\label{fig:teaser}
\end{center}
}]

\begin{abstract}
During 3D reconstruction, it is often the case that people cannot scan each individual object from all views, resulting in missing geometry in the captured scan. This missing geometry can be fundamentally limiting for many applications, e.g., a robot needs to know the unseen geometry to perform a precise grasp on an object. Thus, we introduce the task of semantic instance completion: from an incomplete RGB-D scan of a scene, we aim to detect the individual object instances and infer their complete object geometry. This will open up new possibilities for interactions with objects in a scene, for instance for virtual or robotic agents. We tackle this problem by introducing \OURS{}, a new data-driven approach that jointly detects object instances and predicts their complete geometry. This enables a semantically meaningful decomposition of a scanned scene into individual, complete 3D objects, including hidden and unobserved object parts. \OURS{} is an end-to-end 3D neural network architecture that leverages joint color and geometry feature learning. 
The fully-convolutional nature of our 3D network enables efficient inference of semantic instance completion for 3D scans at scale of large indoor environments in a single forward pass. 
We show that predicting complete object geometry improves both 3D detection and instance segmentation performance.
We evaluate on both real and synthetic scan benchmark data for the new task, where we outperform state-of-the-art approaches by over $15$ in mAP@0.5 on ScanNet, and over 18 in mAP@0.5 on SUNCG.
\end{abstract}

\section{Introduction}
Understanding 3D environments is fundamental to many tasks spanning computer vision, graphics, and robotics. 
In particular, in order to effectively navigate, and moreover interact with an environment, an understanding of the geometry of a scene and the objects it comprises of is essential. 
This is in contrast to the partial nature of reconstructed RGB-D scans; e.g., due to sensor occlusions.
For instance, for a robot exploring an environment, it needs to infer where objects are as well as what lies behind the objects it sees in order to efficiently navigate or perform tasks like grasping.
That is, it needs not only instance-level knowledge of objects in the scene, but to also estimate the missing geometry of these objects.
Additionally, for content creation or mixed reality applications, captured scenes must be decomposable into their complete object components, in order to enable applications such as scene editing or virtual-real object interactions; i.e., it is often insufficient to segment object instances only for observed regions.

Thus, we aim to address this task of ``seeing behind objects,'' which we refer to as \emph{ semantic instance completion}: predicting object detection as well as instance-level completion for an input partial 3D scan of a scene. 
Previous approaches have addressed these tasks independently: 3D instance segmentation segments object instances from the visible surface of a partial scan~\cite{wang2018sgpn,hou20193dsis,yi2018gspn,yang2019learning,lahoud20193d,narita2019panopticfusion,liu2016ssd,elich20193d}, and 3D scan completion approaches predict the full scene geometry~\cite{song2017ssc,dai2018scancomplete}, but lack the notion of individual objects.
In contrast, our approach focuses on the instance level, as knowledge of instances is essential towards enabling interaction with the objects in an environment.

In addition, the task of semantic instance completion is not only important towards enabling object-level understanding and interaction with 3D environments, but we also show that the prediction of complete object geometry informs the task of semantic instance segmentation. Thus, in order to address the task of semantic instance completion, we propose to consider instance detection and object completion in an end-to-end, fully differentiable fashion. 

From an input RGB-D scan of a scene, our \OURS{} model sees behind objects to predict each object's complete geometry. First, object bounding boxes are detected and regressed, followed by object classification and then a prediction of complete object geometry. 
Our approach leverages a unified backbone from which instance detection and object completion are predicted, enabling information to flow from completion to detection. We incorporate features from both color image and 3D geometry of a scanned scene, as well as a fully-convolutional design in order to effectively predict the complete object decomposition of varying-sized scenes.
To address the task of semantic instance completion for real-world scans, where ground truth complete geometry is not readily available, we further introduce a new semantic instance completion benchmark for ScanNet~\cite{dai2017scannet}, leveraging the Scan2CAD~\cite{avetisyan2019scan2cad} annotations to evaluate semantic instance completion (and semantic instance segmentation).

In summary, we present a fully-convolutional, end-to-end 3D CNN formulation to predict 3D instance completion that outperforms state-of-the-art, decoupled approaches to semantic instance completion by 15.8 in mAP@0.5 on real-world scan data, and 18.5 in mAP@0.5 on synthetic data:
\begin{itemize} \itemsep0em 
    \item We introduce the task of \emph{semantic instance completion} for 3D scans;
    \item we propose a novel, end-to-end 3D convolutional network which predicts 3D semantic instance completion as object bounding boxes, class labels, and complete object geometry,
    \item and we show that semantic instance completion task can benefit semantic instance segmentation and detection performance.
\end{itemize}
\section{Related Work}

\paragraph{Object Detection and Instance Segmentation}

Recent advances in convolutional neural networks have now begun to drive impressive progress in object detection and instance segmentation for 2D images~\cite{girshick2015fast,ren2015faster,liu2016ssd,redmon2016you,lin2017feature,he2017mask,lin2018focal}. Combined with the increasing availability of synthetic and real-world 3D data~\cite{dai2017scannet,song2017ssc,chang2017matterport3d}, we are now seeing more advances in object detection~\cite{song2014sliding,song2015deep,qi2017frustum, qi2019deep} for 3D.
Sliding Shapes~\cite{song2014sliding} predicted 3D object bounding boxes from a depth image, designing handcrafted features to detect objects in a sliding window fashion. Deep Sliding Shapes~\cite{song2015deep} then extended this approach to leverage learned features for object detection in a single RGB-D frame. Frustum PointNet~\cite{qi2017frustum} tackles the problem of object detection for an RGB-D frame by first detecting object in the 2D image before projecting the detected boxes into 3D to produce final refined box predictions. VoteNet~\cite{qi2019deep} propose a reformulation of Hough voting in the context of deep learning through an end-to-end differentiable architecture for 3D detection purpose.

Recently, several approaches have been introduced to perform 3D instance segmentation, applicable to single or multi-frame RGB-D input. Wang et al.~\cite{wang2018sgpn} introduced SGPN to operate on point clouds by clustering semantic segmentation predictions. Li et al.~\cite{yi2018gspn} leverages an object proposal-based approach to predict instance segmentation for a point cloud. Simultaneously, Hou et al.~\cite{hou20193dsis} presented an approach leveraging joint color-geometry feature learning for detection and instance segmentation on volumetric data. Lahoud et al.~\cite{lahoud20193d} proposes to use multi-task losses to predict instance segmentation. Yang et al.~\cite{yang2019learning} and Liu et al.~\cite{liu2019masc} both use bottom-up methods to predict instance segmentation for a point cloud. Our approach also leverages an anchor-based object proposal mechanism for detection, but we leverage object completion to predict instance completion, as well as show that completing object-level geometry can improve detection and instance segmentation performance on volumetric data.

\paragraph{3D Scan Completion}
Scan completion of 3D shapes has been a long-studied problem in geometry processing, particularly for cleaning up broken mesh models.
In this context, traditional methods have largely focused on filling small holes by locally fitting geometric primitives, or through continuous energy minimization~\cite{sorkine2004least,nealen2006laplacian,zhao2007robust}.
Surface reconstruction approaches on point cloud inputs~\cite{kazhdan2006poisson,kazhdan2013screened} can also be applied in this fashion to locally optimize for missing surfaces.
Other shape completion approaches leverage priors such as symmetry and structural priors~\cite{thrun2005shape,mitra2006partial,pauly2008discovering,sipiran2014approximate,speciale2016symmetry},  or CAD model retrieval~\cite{nan2012search,shao2012interactive,kim2012acquiring,li2015database,shi2016data} to predict the scan completion.

Recently, methods leveraging generative deep learning have been developed to predict the complete geometry of 3D shapes~\cite{wu20153d,dai2017complete,han2017complete,hane2017hierarchical}.
Song et al.~\cite{song2017ssc} extended beyond shapes to predicting the voxel occupancy for a single depth frame leveraging the geometric occupancy prediction to achieve improved 3D semantic segmentation.
Recently, Dai et al.~\cite{dai2018scancomplete} presented a first approach for data-driven scan completion of full 3D scenes, leveraging a fully-convolutional, autoregressive approach to predict complete geometry along with 3D semantic segmentation.
Both Song et al.~\cite{song2017ssc} and Dai et al.~\cite{dai2018scancomplete} show that inferring the complete scan geometry can improve 3D semantic segmentation.
With our approach for 3D semantic instance completion, this task not only enables new applications requiring instance-based knowledge of a scene (e.g., virtual or robotic interactions with objects in a scene), but we also show that instance segmentation can benefit from instance completion.
\section{Method Overview}
\begin{figure*}[tp]
\begin{center}
\includegraphics[width=1.0\linewidth]{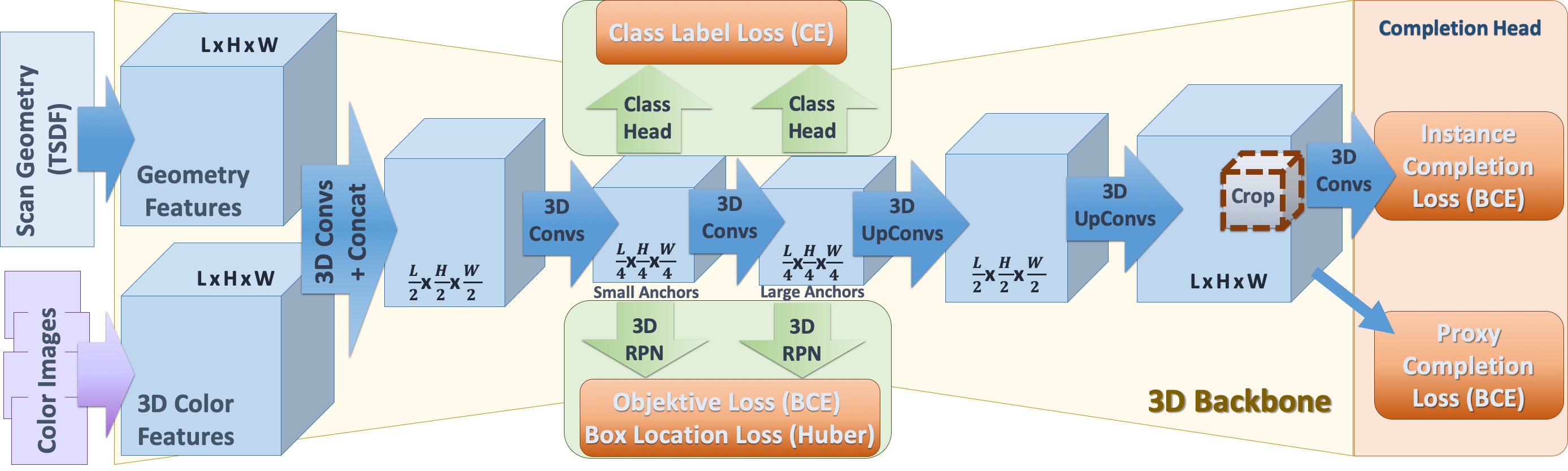}
\end{center}
   \caption{Our \OURS{} network architecture takes an RGB-D scan as input. Color images are processed with 2D convolutions to spatially compress the information before back-projecting into 3D, to be merged with the 3D geometry features of the scan (following~\cite{dai20183dmv,hou20193dsis}). These joint features are used for object detection (as 3D bounding boxes and class labels) followed by per-instance geometric completion, for the task of semantic instance completion. In contrast to \cite{hou20193dsis}, which leverages separate backbones for detection and instance segmentation, our network maintains one unified backbone for both detection and completion head, allowing the completion task to directly inform the detection parameters.}
\label{fig:architecture}
\end{figure*}

Our network takes as input an RGB-D scan, and learns to join together features from both the color images as well as the 3D geometry to inform the semantic instance completion. The architecture is shown in Fig.~\ref{fig:architecture}.

The input 3D scan is encoded as a truncated signed distance field (TSDF) in a volumetric grid.
To combine this with color information from the RGB images, we first extract 2D features using 2D convolutional layers on the RGB images, which are then back-projected into a 3D volumetric grid, and subsequently merged with geometric features extracted from the geometry.
The joint features are then fed into an encoder-decoder backbone, which leverages a series of 3D residual blocks to learn the representation for the task of semantic instance completion. Objects are detected through anchor proposal and bounding box regression; these predicted object boxes are then used to crop and extract features from the backbone encoder to predict the object class label as well as the complete object geometry for each detected object as per-voxel occupancies. 

We adopt in total five losses to supervise the learning process illustrated in Fig.~\ref{fig:architecture}. Detection contains three losses: (1) objectness using binary cross entropy to indicate that there is an object, (2) box location using a Huber loss to regress the 3D bounding box locations, and (3) classification of the class label loss using cross entropy. Following detection, the completion head contains two losses: per-instance completion loss using binary cross entropy to predict per-voxel occupancies, and a proxy completion loss using binary cross entropy to classify the surface voxels belonging to all objects in the scene.

Our method operates on a unified backbone for detection followed by instance completion, enabling object completion to inform the object detection process; this results in effective 3D detection as well as instance completion.
Its fully-convolutional nature enables us to train on cropped chunks of 3D scans but test on a whole scene in a single forward pass, resulting in an efficient decomposition of a scan into a set of complete objects.
\section{Network Architecture}
From an RGB-D scan input, our network operates on the scan's reconstructed geometry, encoded as a TSDF in a volumetric grid, as well as the color images.
To jointly learn from both color and geometry, color features are first extracted in 2D with a 2D semantic segmentation network~\cite{paszke2016enet}, and then back-projected into 3D to be combined with the TSDF features, similar to~\cite{dai20183dmv,hou20193dsis}.
This enables complementary semantic features to be learned from both data modalities. These features are then input to the backbone of our network, which is structured in an encoder-decoder style.

The encoder-decoder backbone is composed of a series of five 3D residual blocks, which generates five volumetric feature maps $\mathbb{F} = \{f_i|i=1 \dots 5\}$. The encoder results in a reduction of spatial dimension by a factor of $4$, and symmetric decoder results in an expansion of spatial dimension by a factor of 4. 
Skip connections link spatially-corresponding encoder and decoder features.
For a more detailed description of the network architecture, we refer to the appendix.
\label{sec:architecture}

\subsection{Color Back-Projection}
As raw color data is often of much higher resolution than 3D geometry, to effectively learn from both color and geometry features, we leverage color information by back-projecting 2D CNN features learned from RGB images to 3D, similar to~\cite{dai20183dmv,hou20193dsis}.
For each voxel location $v_i = (x, y, z)$ in the 3D volumetric grid, we find its pixel location $p_i = (x, y)$ in 2D views by camera intrinsic and extrinsic matrices. We assign the voxel feature at location $v_i$ with the learned 2D CNN feature vector at $p_i$. 
To handle multiple image observations of the same voxel $v_i$, we apply element-wise view pooling; this also allows our approach to handle a varying number of input images. 
Note that this back-projection is differentiable, allowing our model to be trained end-to-end and benefit from both RGB and geometric signal.

\subsection{Object Detection}
For object detection, we predict the bounding box of each detected object as well as the class label.
To inform the detection, features are extracted from feature maps $F_2$ and $F_3$ of the backbone encoder. 
We define two set of anchors on these two features maps, $A_s=\{a_i|i=1 \dots N_s\}$ and $A_b=\{a_i|i=1 \dots N_b\}$ representing `small' and `large' anchors for the earlier $F_2$ and later $F_3$, respectively, so that the larger anchors are associated with the feature map of larger receptive field.
These anchors $A_s \cup A_b$ are selected through a k-means clustering of the ground truth 3D bounding boxes.
For our experiments, we use $N_s + N_b=9$. From these $N_s + N_b$ clusters, $A_b$ are those with any axis $>1.125$m, and the rest are in $A_s$.

The two features maps $F_2$ and $F_3$ are then processed by a 3D region proposal to regress the 3D object bounding boxes.
The 3D region proposal first employs a $1\times 1\times 1$ convolution layer to output objectness scores for each potential anchor, producing an objectness feature map with $2(N_s + N_b)$ channels for the positive and negative objectness probabilities.
Another $1\times 1\times 1$ convolution layer is used to predict the 3D bounding box locations as $6$-dimensional offsets from the anchors; we then apply a non-maximum suppression based on the objectness scores.
We use a Huber loss on the log ratios of the offsets to the anchor sizes to regress the final bounding box predictions:
\begin{equation*} 
\Delta_x = \frac{\mu - \mu_{anchor}}{\phi_{anchor}}  \;\;\;\;\;\;
\Delta_w = \ln(\frac{\phi}{\phi_{anchor}})  
\end{equation*}
where $\mu$ is the box center point and $\phi$ is the box width.
The final bounding box loss is then:
\[
    L_{\Delta}= 
\begin{cases}
    \frac{1}{2} \Delta^{2},& \text{if $|\Delta|$ } \leq 2\\
    |\Delta|,              & \text{otherwise.}
\end{cases}
\]

Using these predicted object bounding boxes, we then predict the object class labels using features cropped from the bounding box locations from $F_2$ and $F_3$.
We use a 3D region of interest pooling layer to unify the sizes of the cropped feature maps to a spatial dimension of $4\times 4\times 4$ to be input to an object classification MLP.

\subsection{Instance Completion}
For each object, we infer its complete geometry by predicting per-voxel occupancies.
Here, we crop features from feature map $F_5$ of the backbone, which has a feature map resolution matching the input spatial resolution, using the predicted object bounding box.
These features are processed through a series of five 3D convolutions which maintain the spatial resolution of their input. 
The complete geometry is then predicted as voxel occupancy using a binary cross entropy loss.

We predict $N_\textrm{classes}$ potential object completions for each class category, and select the final prediction based on the predicted object class. 
We define ground truth bounding boxes $b_i$ and masks $m_i$ as  ${\gamma} = \{(b_i, m_i)|i=1 \dots N_b\}$. 
Further, we define  predicted bounding boxes $\hat{b_i}$ along with predicted masks $\hat{m_i}$ as $\hat{\gamma} = \{(\hat{b_i}, \hat{m_i})|i=1 \dots \hat{N_b}\}$.
During training, we only train on predicted bounding boxes that overlap with the ground truth bounding boxes:
\begin{align*}
\Omega = \{(&\hat{b_i}, \hat{m_i}, b_i, m_i)\;|\; \text{IoU}(\hat{b_i}, b_i) \geq 0.5,\\
&\forall (\hat{b_i}, \hat{m_i}) \in \hat{\gamma}, \forall (b_i, m_i) \in \gamma \}
\end{align*}

We can then define the instance completion loss for each associated pair in $\Omega$:
\begin{align*}
L_{\text{compl}} & = \frac{1}{|\Omega|} \sum_{\Omega} \text{BCE}(\text{sigmoid}(\hat{m_i}), m_i'), \\
m_i'(v) & = \begin{cases}
    m_i(v) & \text{if}\; v \in \hat{b_i}\cap b_i\\
    0      &  \text{otherwise.}
\end{cases}
\end{align*}

We further introduce a global geometric completion loss on entire scene level that serves as an intermediate proxy. 
To this end, we use feature map $F_5$ as input to a binary cross entropy loss whose target is the composition of all complete object instances of the scene:
\begin{equation*}
L_{\text{geometry}} = \text{BCE}(\text{sigmoid}(F_5), \cup_{(b_i, m_i) \in \gamma}).
\end{equation*}

Our intuition is to obtain a strong gradient during training by adding this additional constraint to each voxel in the last feature map $F_5$. 
We find that this global geometric completion loss further helps the final instance completion performance; see Sec~\ref{sec:results}.
\begin{table*}[tp!]
    \centering
     \resizebox{0.9\textwidth}{!}{
     \begin{tabular}{l|cccccccc||c}\specialrule{1.3pt}{0.0pt}{0.1pt}
     & display & table & bathtub & trashbin & sofa & chair & cabinet & bookshelf & \textbf{avg}\\ \hline
Scene Completion + Instance Segmentation & 1.65   &0.64  & 4.55   & 11.25& 9.09& 9.09 & 0.18   & 5.45 & 5.24\\ 
Instance Segmentation + Shape Completion & 2.27   &3.90  & 1.14 & 1.68 & 14.86& 9.93 & 7.11 & 3.03 & 5.49\\ \hline
Ours -- \OURS{} (no color)  & 13.16   & 11.28 & 13.64   & 18.19   & 24.79  & 15.87 & 8.60  & 10.60 & 14.52\\ 
Ours -- \OURS{} (no proxy)  & 21.94   &7.63   & 12.55   & 28.24   & 20.38  & 22.58 & 13.42 & 9.51  & 17.03\\ 
Ours -- \OURS{}             & {\bf 26.86}   & {\bf 13.21}  & {\bf 22.31}   & {\bf 28.93}   & {\bf 29.41}  & {\bf 23.64} & {\bf 15.35} & {\bf 14.48} & {\bf 21.77}\\ 
    \specialrule{1.3pt}{0.1pt}{0pt}
    \end{tabular}
    }
    \caption{3D Semantic Instance Completion on ScanNet~\cite{dai2017scannet} scans with Scan2CAD~\cite{avetisyan2019scan2cad} targets at mAP@0.5.
    Our end-to-end formulation achieves significantly better performance than alternative, decoupled approaches that first use state-of-the-art scan completion~\cite{dai2018scancomplete} and then instance segmentation~\cite{hou20193dsis} method or first instance segmentation~\cite{hou20193dsis} and then shape completion~\cite{dai2017complete}.}
    \label{tab:scannet_completion}
\end{table*}

\section{Network Training}
\label{sec:training}

\subsection{Data}
The input 3D scans are represented as truncated signed distance fields (TSDFs) encoded in volumetric grids.
The TSDFs are generated through volumetric fusion~\cite{curless1996volumetric} during the 3D reconstruction process.
For all our experiments, we used a voxel size of $\approx 4.7$cm and truncation of $3$ voxels.
We also input the color images of the RGB-D scan, which we project to the 3D grid using their camera poses.
We train our model on both synthetic and real scans, computing $9$ anchors through $k$-means clustering; for real-world ScanNet~\cite{dai2017scannet} data, this results in $4$ small anchors and $5$ large anchors, and for synthetic SUNCG~\cite{song2017ssc} data, this results in $3$ small anchors and $6$ large anchors.

At test time, we leverage the fully-convolutional design to input the full scan of a scene along with its color images.
During training, we use random $96\times 48\times 96$ crops ($4.5\times 2.25\times 4.5$ meters) of the scanned scenes, along with a greedy selection of $\leq 5$ images covering the most object geometry in the crop. 
Only objects with $50\%$ of their complete geometry inside the crop are considered.

\subsection{Optimization}
We train our model jointly, end-to-end from scratch.
We use an SGD optimizer with batch size 64 for object proposals and 16 for object classification, and all positive bounding box predictions ($>0.5$ IoU with ground truth box) for object completion.
We use a learning rate of 0.005, which is decayed by a factor of 0.1 every 100k steps.
We train our model for 200k steps ($\approx$ 60 hours) to convergence, on a single Nvidia GTX 1080Ti.
Additionally, we augment the data for training the object completion using ground truth bounding boxes and classification in addition to predicted object detection.

\begin{figure*}[h!]
\begin{center}
   \includegraphics[width=1.0\linewidth]{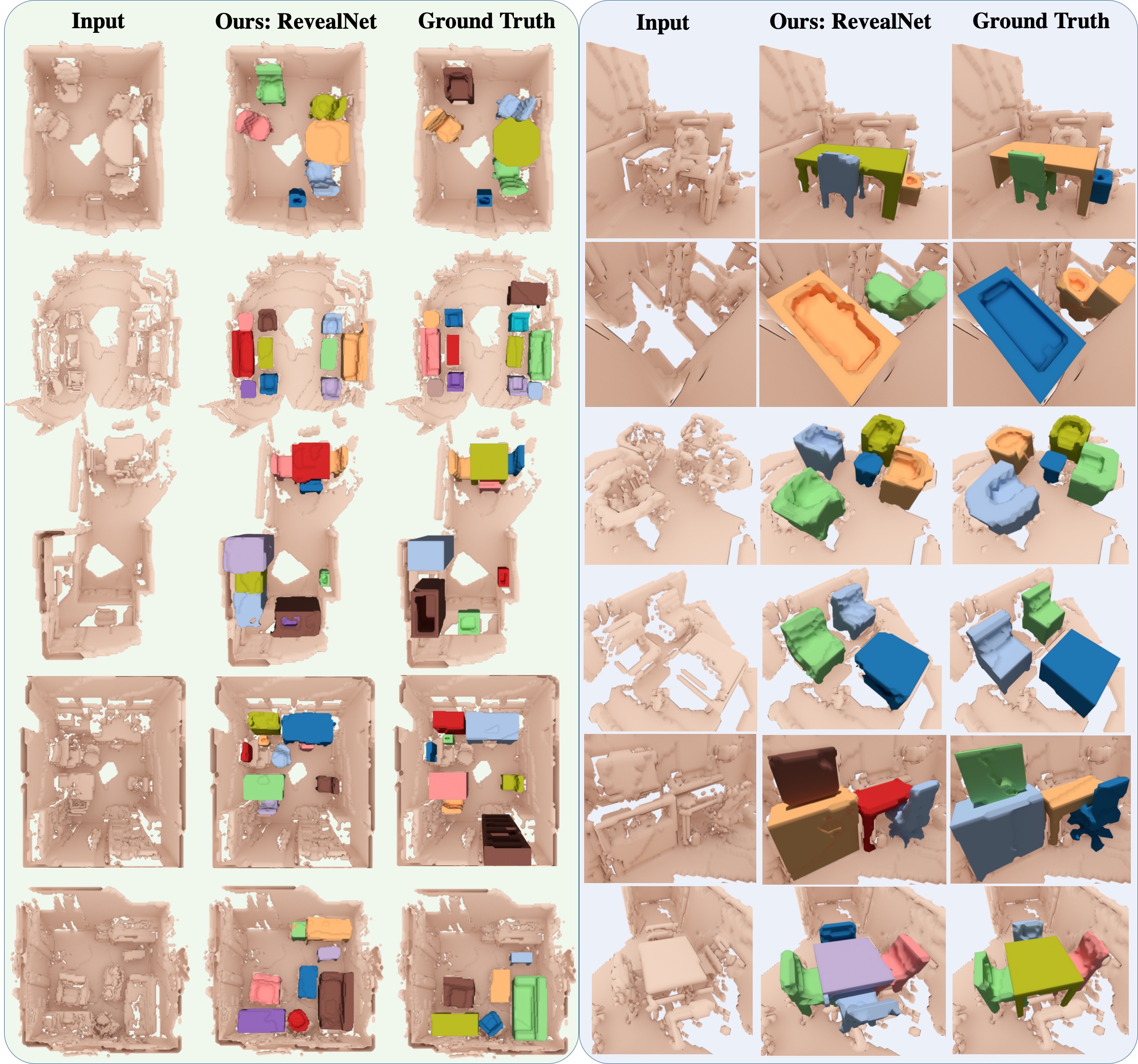}
    \end{center}
     \vspace{-0.2cm}
   \caption{
    Qualitative results on real-world ScanNet~\cite{dai2017scannet} scenes with Scan2CAD~\cite{avetisyan2019scan2cad} targets. Close-ups are shown on the right.
    Note that different colors denote distinct object instances in the visualization.
   Our approach effectively predicts complete individual object geometry, including missing structural components (e.g., missing chair legs), across varying degrees of partialness in input scan observations.
    }
 \vspace{-0.3cm}
\label{fig:results_comparison_scannet}
\end{figure*}

\begin{table*}[h!]
    \centering
     \resizebox{\textwidth}{!}{
     \setlength{\tabcolsep}{2pt}
     \begin{tabular}{l|ccccccccccccccccccccccc||c}
     \specialrule{1.3pt}{0.0pt}{0.1pt}
     & cab & bed & chair & sofa & tabl & door & wind & bkshf & cntr & desk & shlf&curt &drsr&mirr&tv&nigh& toil & sink & lamp&bath & ostr & ofurn&oprop& \textbf{avg}\\ \hline
     SC + IS & 3.0&0.6&19.5&0.8&18.1&\textbf{15.9}&0.00&0.0&1.0&2.3&3.0&0.0&0.5&0.0&9.2&10.4&23.9&3.4&9.1&0.0&0.0&0.0&9.1&5.5 \\
     IS + SC & 0.3 & 0.0  & 7.4 & 0.4 & 3.0 & 9.1 & 0.0 & 0.0 & 0.2 &0.0&0.0&0.0&2.3&0.0&3.0&0.0&2.6&0.0&1.8&0.0&0.0&0.0&4.6&1.5\\ \hline
     no color & \textbf{19.05}&41.8&38.2&11.9&23.9&9.1&0.0&0.0&2.5&\textbf{21.6}&9.1&0.0&\textbf{12.6}&\textbf{4.6}&49.4&33.8&63.4&\textbf{36.9}&\textbf{38.8}&14.7&15.9&0.0&\textbf{23.8}&20.5 \\
     no proxy & 12.9&46.1&\textbf{39.4}&26.8&\textbf{30.3}&1.0&15.9&0.0&9.1&18.2&3.4&0.0&1.1&0.0&43.6&\textbf{34.0}&\textbf{69.1}&32.4&29.6&\textbf{31.1}&14.6&0.0&23.3&20.9 \\
     Ours &14.7&\textbf{58.3}&38.2&\textbf{28.8}&29.5&0.0&\textbf{15.9}&\textbf{54.6}&\textbf{9.1}&12.1&\textbf{9.1}&\textbf{0.0}&6.2&0.0&\textbf{49.4}&33.5&61.2&34.5&29.5&27.1&\textbf{16.4}&\textbf{0.0}&23.5&\textbf{24.0} \\
    \specialrule{1.3pt}{0.1pt}{0pt}
    \end{tabular}
    }
     \vspace{-0.1cm}
    \caption{3D Semantic Instance Completion on synthetic SUNCG~\cite{song2017ssc} scans at mAP@0.5.
    Our semantic instance completion approach achieves significantly better performance than alternative approaches with decoupled state-of-the-art scan completion (SC)~\cite{dai2018scancomplete} followed by instance segmentation (IS)~\cite{hou20193dsis}, as well as instance segmentation followed by shape completion~\cite{dai2017complete}.
    We additionally evaluate our approach without color input (no color) and without a completion proxy loss on the network backbone (no proxy).}
    \label{tab:suncg_completion}

\end{table*}

\section{Results}
\label{sec:results}

\begin{figure*}[h!]
\begin{center}
   \includegraphics[width=1.0\linewidth]{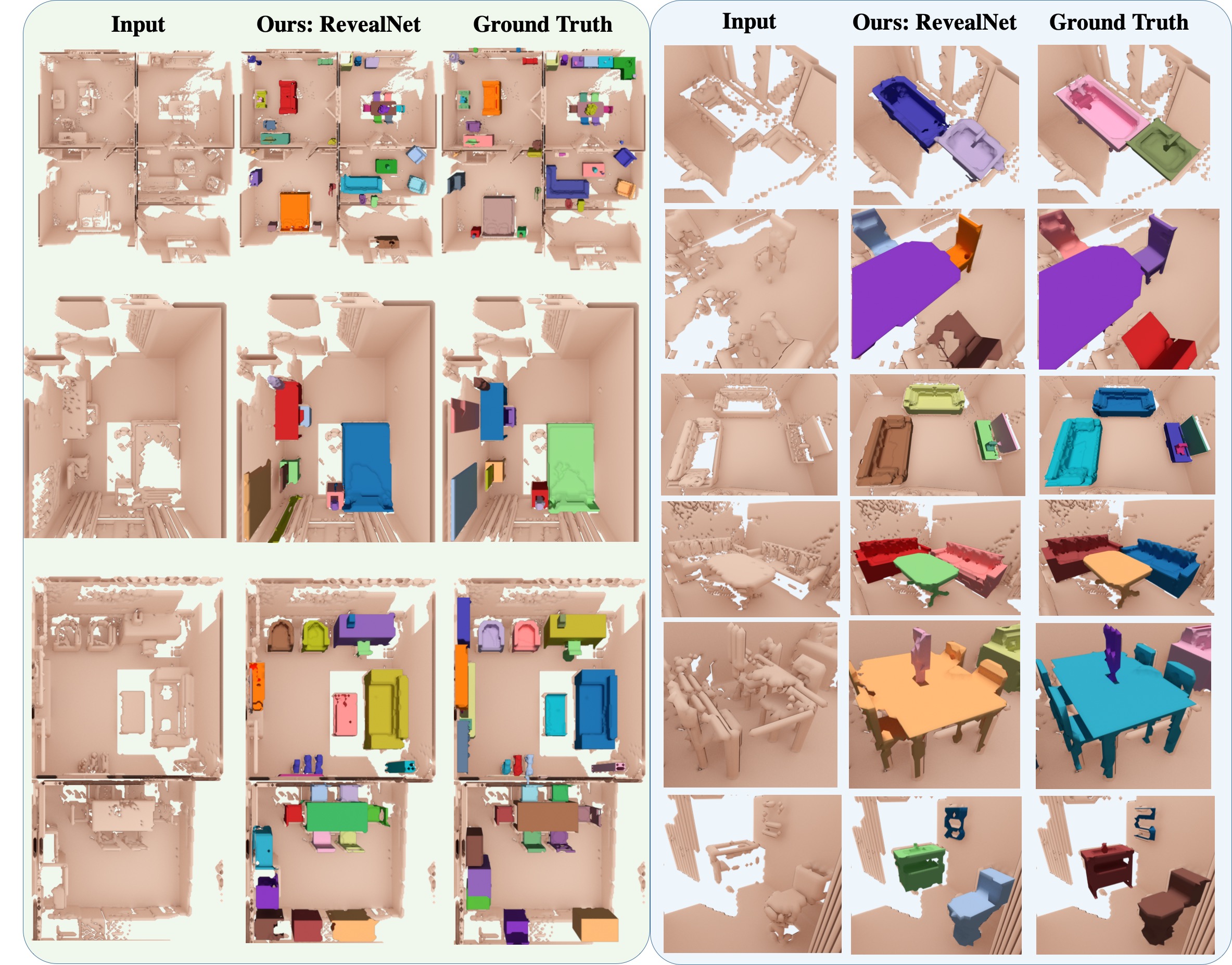}
    \end{center}
  \vspace{-0.4cm}
   \caption{
    Qualitative results on SUNCG dataset~\cite{song2017ssc} (left: full scans, right: close-ups). We sample RGB-D images to reconstruct incomplete 3D scans from random camera trajectories inside SUNCG scenes. Note that different colors denote distinct object instances in the visualization.
    }
    \vspace{-0.3cm}
\label{fig:results_comparison_suncg}
\end{figure*}

We evaluate our approach on semantic instance completion performance on synthetic scans of SUNCG~\cite{song2017ssc} scenes as well as on real-world ScanNet~\cite{dai2017scannet} scans, where we obtain ground truth object locations and geometry from CAD models aligned to ScanNet provided by Scan2CAD~\cite{avetisyan2019scan2cad}.
To evaluate semantic instance completion, we use a mean average precision metric on the complete masks (at IoU $0.5$).
Qualitative results are shown in Figs.~\ref{fig:results_comparison_scannet} and \ref{fig:results_comparison_suncg}.

\begin{table}[tp!]
    \centering
    \small
     \begin{tabular}{l|c|c}\specialrule{1.3pt}{0.0pt}{0.1pt}
         &3D Detection & Instance Segmentation \\ \hline
3D-SIS~\cite{hou20193dsis} & 25.70 & 20.78\\ \hline
Ours (no compl) & 31.93& 24.49\\ 
Ours (no color) & 29.29 & 23.55\\ 
Ours (no proxy) & 31.52 & 25.92\\ 
Ours  & {\bf36.39}  & {\bf 30.52}\\ 
    \specialrule{1.3pt}{0.1pt}{0pt}
    \end{tabular}

    \caption{3D Detection and Instance Segmentation on ScanNet~\cite{dai2017scannet} scans with Scan2CAD~\cite{avetisyan2019scan2cad} annotations at mAP@0.5.
    We evaluate our instance completion approach on the task of instance segmentation and detection to justify our contribution that instance completion task helps instance segmentation and detection. We evaluate our approach without completion (no compl), without color input (no color), and without a completion proxy loss on the network backbone (no proxy).
    Predicting instance completion notably increases performance of predicting both instance segmentation and detection (Ours vs. no compl).  We additionally compare against 3D-SIS~\cite{hou20193dsis}, a state-of-the-art approach for both 3D detection and instance segmentation on 3D dense volumetric data (the representation we use).}
    
    \label{tab:scannet_instance}

\end{table}

\begin{table}[tp!]
    \centering
    \small
     \begin{tabular}{l|c|c}\specialrule{1.3pt}{0.0pt}{0.1pt}
     & 3D Detection & Instance Segmentation \\ \hline
3D-SIS~\cite{hou20193dsis} &24.70 &20.61\\ \hline
Ours (no compl) &29.80 &23.86\\ 
Ours (no color) & 31.75&31.59\\ 
Ours (no proxy) & 34.05&32.59\\ 
Ours &\textbf{37.81}&\textbf{36.28}\\ 
    \specialrule{1.3pt}{0.1pt}{0pt}
    \end{tabular}
    
    \caption{3D Detection and Instance Segmentation on synthetic SUNCG~\cite{song2017ssc} scans at mAP@0.5. To demonstrate the benefits of instance completion task for instance segmentation and 3D detection, we evaluate our semantic instance completion approach on the task of instance segmentation and 3D detection.
    Predicting instance completion notably benefits 3D detection and instance segmentation (Ours vs. no compl). 
    }
   \vspace{-0.5cm}
    \label{tab:suncg_instance}
\end{table}

\paragraph{Comparison to state-of-the-art approaches for semantic instance completion.}
Tables~\ref{tab:scannet_completion} and \ref{tab:suncg_completion} evaluate our method against state of the art for the task of semantic instance completion on our real and synthetic scans, respectively.
Qualitative comparisons on ScanNet scans~\cite{dai2017scannet} with Scan2CAD~\cite{avetisyan2019scan2cad} targets (which provide ground truth for complete object geometry) are shown in Fig.~\ref{fig:results_comparison_scannet}.
We compare to state-of-the-art 3D instance segmentation and scan completion approaches used sequentially; that is, first applying a 3D instance segmentation approach followed by a shape completion method on the predicted instance segmentation, as well as first applying a scene completion approach to the input partial scan, followed by a 3D instance segmentation method.
For 3D instance segmentation, we evaluate 3D-SIS~\cite{hou20193dsis}, which achieves state-of-the-art performance on a dense volumetric grid representation (the representation we use), and for scan completion we evaluate the 3D-EPN~\cite{dai2017complete} shape completion approach and ScanComplete~\cite{dai2018scancomplete} scene completion approach.
Our end-to-end approach for semantic instance completion results in significantly improved performance due to information flow from instance completion to object detection.
For instance, this allows our instance completion to more easily adapt to some inaccuracies in detection, which strongly hinders a decoupled approach.
Note that the ScanComplete model applied on ScanNet data is trained on synthetic data, due to the lack of complete ground truth scene data (Scan2CAD provides only object ground truth) for real-world scans.
\vspace{-0.2cm}

\paragraph{Does instance completion help instance detection and segmentation?}
We can also evaluate our semantic instance completion predictions on the task of semantic instance segmentation by taking the intersection between the predicted complete mask and the input partial scan geometry to be the predicted instance segmentation mask.
We show that predicting instance completion helps instance segmentation, evaluating our method on 3D semantic instance segmentation with and without completion, on ScanNet~\cite{dai2017scannet} and SUNCG~\cite{song2017ssc} scans in Tables~\ref{tab:scannet_instance} and \ref{tab:suncg_instance}, as well as 3D-SIS~\cite{hou20193dsis}, an approach jointly predicts 3D detection and instance segmentation, which also operates on dense volumetric data, achieving state-of-the-art performance on this representation. 
We find that predicting instance completion significantly benefits instance segmentation, due to a more unified understanding of object geometric structures. 

Additionally, we evaluate the effect on 3D detection in Tables~\ref{tab:scannet_instance} and \ref{tab:suncg_instance}; predicting instance completion also significantly improves 3D detection performance. Note that in contrast to 3D-SIS~\cite{hou20193dsis} which uses separate backbones for detection and instance segmentation, our unified backbone helps 3D mask information (complete or non-complete) propagate through detection parameters to improve 3D detection performance.

\paragraph{What is the effect of a global completion proxy?}
In Tables~\ref{tab:scannet_completion} and \ref{tab:suncg_completion}, we demonstrate the impact of the geometric completion proxy loss; here, we see that this loss improves the semantic instance completion performance on both real and synthetic data.
In Tables~\ref{tab:scannet_instance} and \ref{tab:suncg_instance}, we can see that it also improves 3D detection and semantic instance segmentation performance.

\paragraph{Can color input help?}
Our approach takes as input the 3D scan geometry as a TSDF as well as the corresponding color images. We evaluate our approach with and without the color input stream; on both real and synthetic scans, the color input notably improves semantic instance completion performance, as shown in Tables~\ref{tab:scannet_completion} and \ref{tab:suncg_completion}.

\section{Limitations}
Our approach shows significant potential in the task of semantic instance completion, but several important limitations still remain.
First, we output a binary mask for the complete object geometry, which can limit the amount of detail represented by the completion; other 3D representations such as distance fields or sparse 3D representations~\cite{SubmanifoldSparseConvNet} could potentially resolve greater geometric detail.
Our approach also uses axis-aligned bounding boxes for object detection; it would be helpful to additionally predict the object orientation.
We also do not consider object movement over time, which contains significant opportunities for semantic instance completion in the context of dynamic environments.

\section{Conclusion}
In this paper, we tackle the problem of ``seeing behind objects'' by predicting the missing geometry of individual objects in RGB-D scans. 
This opens up many possibilities for complex interactions with objects in 3D, for instance for efficient navigation or robotic grasping. 
To this end, we introduced the new task of semantic instance completion along with \OURS{}, a new 3D CNN-based approach to jointly detect objects and predict their complete geometry.
Our proposed 3D CNN learns from both color and geometry features to detect and classify objects, then predicts the voxel occupancy for the complete geometry of the object in an end-to-end fashion, which can be run on a full 3D scan in a single forward pass.
On both real and synthetic scan data, we significantly outperform state-of-the-art approaches for semantic instance completion.
We believe that our approach makes an important step towards higher-level scene understanding and helps to enable object-based interactions and understanding of scenes, which we hope will open up new research avenues.


\section*{Acknowledgments}
This work was supported by the ZD.B, a Google Research Grant, an Nvidia Professor Partnership, a TUM Foundation Fellowship, a TUM-IAS Rudolf M{\"o}{\ss}bauer Fellowship, and the ERC Starting Grant \emph{Scan2CAD (804724)}.

{\small
\bibliographystyle{ieee_fullname}
\bibliography{egbib}

\begin{thebibliography}{10}\itemsep=-1pt

\bibitem{avetisyan2019scan2cad}
Armen Avetisyan, Manuel Dahnert, Angela Dai, Manolis Savva, Angel~X. Chang, and
  Matthias Nie{\ss}ner.
\newblock Scan2cad: Learning cad model alignment in rgb-d scans.
\newblock In {\em Proc. Computer Vision and Pattern Recognition (CVPR), IEEE},
  2019.

\bibitem{avetisyan2019end2end}
Armen Avetisyan, Angela Dai, and Matthias Nie{\ss}ner.
\newblock End-to-end cad model retrieval and 9dof alignment in 3d scans.
\newblock In {\em ICCV 2019}, 2019.

\bibitem{chang2017matterport3d}
Angel Chang, Angela Dai, Thomas Funkhouser, Maciej Halber, Matthias Niessner,
  Manolis Savva, Shuran Song, Andy Zeng, and Yinda Zhang.
\newblock {Matterport3D}: Learning from {RGB-D} data in indoor environments.
\newblock {\em International Conference on 3D Vision (3DV)}, 2017.

\bibitem{curless1996volumetric}
Brian Curless and Marc Levoy.
\newblock A volumetric method for building complex models from range images.
\newblock In {\em Proceedings of the 23rd annual conference on Computer
  graphics and interactive techniques}, pages 303--312. ACM, 1996.

\bibitem{dai2017scannet}
Angela Dai, Angel~X. Chang, Manolis Savva, Maciej Halber, Thomas Funkhouser,
  and Matthias Nie{\ss}ner.
\newblock Scannet: Richly-annotated 3d reconstructions of indoor scenes.
\newblock In {\em Proc. Computer Vision and Pattern Recognition (CVPR), IEEE},
  2017.

\bibitem{dai20183dmv}
Angela Dai and Matthias Nie{\ss}ner.
\newblock 3dmv: Joint 3d-multi-view prediction for 3d semantic scene
  segmentation.
\newblock In {\em Proceedings of the European Conference on Computer Vision
  ({ECCV})}, 2018.

\bibitem{dai2017complete}
Angela Dai, Charles~Ruizhongtai Qi, and Matthias Nie{\ss}ner.
\newblock Shape completion using 3d-encoder-predictor cnns and shape synthesis.
\newblock In {\em Proc. Computer Vision and Pattern Recognition (CVPR), IEEE},
  2017.

\bibitem{dai2018scancomplete}
Angela Dai, Daniel Ritchie, Martin Bokeloh, Scott Reed, J{\"u}rgen Sturm, and
  Matthias Nie{\ss}ner.
\newblock Scancomplete: Large-scale scene completion and semantic segmentation
  for 3d scans.
\newblock In {\em Proc. Computer Vision and Pattern Recognition (CVPR), IEEE},
  2018.

\bibitem{elich20193d}
Cathrin Elich, Francis Engelmann, Jonas Schult, Theodora Kontogianni, and
  Bastian Leibe.
\newblock 3d-bevis: Birds-eye-view instance segmentation.
\newblock {\em arXiv preprint arXiv:1904.02199}, 2019.

\bibitem{girshick2015fast}
Ross Girshick.
\newblock Fast r-cnn.
\newblock In {\em Proceedings of the IEEE international conference on computer
  vision}, pages 1440--1448, 2015.

\bibitem{SubmanifoldSparseConvNet}
Benjamin Graham and Laurens van~der Maaten.
\newblock Submanifold sparse convolutional networks.
\newblock {\em arXiv preprint arXiv:1706.01307}, 2017.

\bibitem{han2017complete}
Xiaoguang Han, Zhen Li, Haibin Huang, Evangelos Kalogerakis, and Yizhou Yu.
\newblock {High Resolution Shape Completion Using Deep Neural Networks for
  Global Structure and Local Geometry Inference}.
\newblock In {\em IEEE International Conference on Computer Vision (ICCV)},
  2017.

\bibitem{hane2017hierarchical}
Christian H{\"a}ne, Shubham Tulsiani, and Jitendra Malik.
\newblock Hierarchical surface prediction for 3d object reconstruction.
\newblock {\em arXiv preprint arXiv:1704.00710}, 2017.

\bibitem{he2017mask}
Kaiming He, Georgia Gkioxari, Piotr Doll{\'a}r, and Ross Girshick.
\newblock Mask r-cnn.
\newblock In {\em Computer Vision (ICCV), 2017 IEEE International Conference
  on}, pages 2980--2988. IEEE, 2017.

\bibitem{hou20193dsis}
Ji Hou, Angela Dai, and Matthias Niessner.
\newblock 3d-sis: 3d semantic instance segmentation of rgb-d scans.
\newblock In {\em The IEEE Conference on Computer Vision and Pattern
  Recognition (CVPR)}, June 2019.

\bibitem{kazhdan2006poisson}
Michael Kazhdan, Matthew Bolitho, and Hugues Hoppe.
\newblock Poisson surface reconstruction.
\newblock In {\em Proceedings of the fourth Eurographics symposium on Geometry
  processing}, volume~7, 2006.

\bibitem{kazhdan2013screened}
Michael Kazhdan and Hugues Hoppe.
\newblock Screened poisson surface reconstruction.
\newblock {\em ACM Transactions on Graphics (TOG)}, 32(3):29, 2013.

\bibitem{kim2012acquiring}
Young~Min Kim, Niloy~J Mitra, Dong-Ming Yan, and Leonidas Guibas.
\newblock Acquiring 3d indoor environments with variability and repetition.
\newblock {\em ACM Transactions on Graphics (TOG)}, 31(6):138, 2012.

\bibitem{lahoud20193d}
Jean Lahoud, Bernard Ghanem, Marc Pollefeys, and Martin~R Oswald.
\newblock 3d instance segmentation via multi-task metric learning.
\newblock {\em arXiv preprint arXiv:1906.08650}, 2019.

\bibitem{li2015database}
Yangyan Li, Angela Dai, Leonidas Guibas, and Matthias Nie{\ss}ner.
\newblock Database-assisted object retrieval for real-time 3d reconstruction.
\newblock In {\em Computer Graphics Forum}, volume~34, pages 435--446. Wiley
  Online Library, 2015.

\bibitem{lin2017feature}
Tsung-Yi Lin, Piotr Doll{\'a}r, Ross~B Girshick, Kaiming He, Bharath Hariharan,
  and Serge~J Belongie.
\newblock Feature pyramid networks for object detection.
\newblock In {\em CVPR}, volume~1, page~4, 2017.

\bibitem{lin2018focal}
Tsung-Yi Lin, Priyal Goyal, Ross Girshick, Kaiming He, and Piotr Doll{\'a}r.
\newblock Focal loss for dense object detection.
\newblock {\em IEEE transactions on pattern analysis and machine intelligence},
  2018.

\bibitem{liu2019masc}
Chen Liu and Yasutaka Furukawa.
\newblock Masc: Multi-scale affinity with sparse convolution for 3d instance
  segmentation.
\newblock {\em arXiv preprint arXiv:1902.04478}, 2019.

\bibitem{liu2016ssd}
Wei Liu, Dragomir Anguelov, Dumitru Erhan, Christian Szegedy, Scott Reed,
  Cheng-Yang Fu, and Alexander~C Berg.
\newblock Ssd: Single shot multibox detector.
\newblock In {\em European conference on computer vision}, pages 21--37.
  Springer, 2016.

\bibitem{mitra2006partial}
Niloy~J Mitra, Leonidas~J Guibas, and Mark Pauly.
\newblock Partial and approximate symmetry detection for 3d geometry.
\newblock In {\em ACM Transactions on Graphics (TOG)}, volume~25, pages
  560--568. ACM, 2006.

\bibitem{nan2012search}
Liangliang Nan, Ke Xie, and Andrei Sharf.
\newblock A search-classify approach for cluttered indoor scene understanding.
\newblock {\em ACM Transactions on Graphics (TOG)}, 31(6):137, 2012.

\bibitem{narita2019panopticfusion}
Gaku Narita, Takashi Seno, Tomoya Ishikawa, and Yohsuke Kaji.
\newblock Panopticfusion: Online volumetric semantic mapping at the level of
  stuff and things.
\newblock {\em arXiv preprint arXiv:1903.01177}, 2019.

\bibitem{nealen2006laplacian}
Andrew Nealen, Takeo Igarashi, Olga Sorkine, and Marc Alexa.
\newblock Laplacian mesh optimization.
\newblock In {\em Proceedings of the 4th international conference on Computer
  graphics and interactive techniques in Australasia and Southeast Asia}, pages
  381--389. ACM, 2006.

\bibitem{paszke2016enet}
Adam Paszke, Abhishek Chaurasia, Sangpil Kim, and Eugenio Culurciello.
\newblock Enet: A deep neural network architecture for real-time semantic
  segmentation.
\newblock {\em arXiv preprint arXiv:1606.02147}, 2016.

\bibitem{pauly2008discovering}
Mark Pauly, Niloy~J Mitra, Johannes Wallner, Helmut Pottmann, and Leonidas~J
  Guibas.
\newblock Discovering structural regularity in 3d geometry.
\newblock In {\em ACM transactions on graphics (TOG)}, volume~27, page~43. ACM,
  2008.

\bibitem{qi2019deep}
Charles~R Qi, Or Litany, Kaiming He, and Leonidas~J Guibas.
\newblock Deep hough voting for 3d object detection in point clouds.
\newblock {\em arXiv preprint arXiv:1904.09664}, 2019.

\bibitem{qi2017frustum}
Charles~R Qi, Wei Liu, Chenxia Wu, Hao Su, and Leonidas~J Guibas.
\newblock Frustum pointnets for 3d object detection from rgb-d data.
\newblock {\em arXiv preprint arXiv:1711.08488}, 2017.

\bibitem{redmon2016you}
Joseph Redmon, Santosh Divvala, Ross Girshick, and Ali Farhadi.
\newblock You only look once: Unified, real-time object detection.
\newblock In {\em Proceedings of the IEEE conference on computer vision and
  pattern recognition}, pages 779--788, 2016.

\bibitem{ren2015faster}
Shaoqing Ren, Kaiming He, Ross Girshick, and Jian Sun.
\newblock Faster r-cnn: Towards real-time object detection with region proposal
  networks.
\newblock In {\em Advances in neural information processing systems}, pages
  91--99, 2015.

\bibitem{shao2012interactive}
Tianjia Shao, Weiwei Xu, Kun Zhou, Jingdong Wang, Dongping Li, and Baining Guo.
\newblock An interactive approach to semantic modeling of indoor scenes with an
  rgbd camera.
\newblock {\em ACM Transactions on Graphics (TOG)}, 31(6):136, 2012.

\bibitem{shi2016data}
Yifei Shi, Pinxin Long, Kai Xu, Hui Huang, and Yueshan Xiong.
\newblock Data-driven contextual modeling for 3d scene understanding.
\newblock {\em Computers \& Graphics}, 55:55--67, 2016.

\bibitem{sipiran2014approximate}
Ivan Sipiran, Robert Gregor, and Tobias Schreck.
\newblock Approximate symmetry detection in partial 3d meshes.
\newblock In {\em Computer Graphics Forum}, volume~33, pages 131--140. Wiley
  Online Library, 2014.

\bibitem{song2014sliding}
Shuran Song and Jianxiong Xiao.
\newblock Sliding shapes for 3d object detection in depth images.
\newblock In {\em European conference on computer vision}, pages 634--651.
  Springer, 2014.

\bibitem{song2015deep}
Shuran Song and Jianxiong Xiao.
\newblock Deep sliding shapes for amodal 3d object detection in rgb-d images.
\newblock {\em arXiv preprint arXiv:1511.02300}, 2015.

\bibitem{song2017ssc}
Shuran Song, Fisher Yu, Andy Zeng, Angel~X Chang, Manolis Savva, and Thomas
  Funkhouser.
\newblock Semantic scene completion from a single depth image.
\newblock {\em Proceedings of 30th IEEE Conference on Computer Vision and
  Pattern Recognition}, 2017.

\bibitem{sorkine2004least}
Olga Sorkine and Daniel Cohen-Or.
\newblock Least-squares meshes.
\newblock In {\em Shape Modeling Applications, 2004. Proceedings}, pages
  191--199. IEEE, 2004.

\bibitem{speciale2016symmetry}
Pablo Speciale, Martin~R Oswald, Andrea Cohen, and Marc Pollefeys.
\newblock A symmetry prior for convex variational 3d reconstruction.
\newblock In {\em European Conference on Computer Vision}, pages 313--328.
  Springer, 2016.

\bibitem{thrun2005shape}
Sebastian Thrun and Ben Wegbreit.
\newblock Shape from symmetry.
\newblock In {\em Tenth IEEE International Conference on Computer Vision
  (ICCV'05) Volume 1}, volume~2, pages 1824--1831. IEEE, 2005.

\bibitem{wang2018sgpn}
Weiyue Wang, Ronald Yu, Qiangui Huang, and Ulrich Neumann.
\newblock Sgpn: Similarity group proposal network for 3d point cloud instance
  segmentation.
\newblock In {\em Proceedings of the IEEE Conference on Computer Vision and
  Pattern Recognition}, pages 2569--2578, 2018.

\bibitem{wu20153d}
Zhirong Wu, Shuran Song, Aditya Khosla, Fisher Yu, Linguang Zhang, Xiaoou Tang,
  and Jianxiong Xiao.
\newblock 3d shapenets: A deep representation for volumetric shapes.
\newblock In {\em Proceedings of the IEEE Conference on Computer Vision and
  Pattern Recognition}, pages 1912--1920, 2015.

\bibitem{yang2019learning}
Bo Yang, Jianan Wang, Ronald Clark, Qingyong Hu, Sen Wang, Andrew Markham, and
  Niki Trigoni.
\newblock Learning object bounding boxes for 3d instance segmentation on point
  clouds.
\newblock {\em arXiv preprint arXiv:1906.01140}, 2019.

\bibitem{yi2018gspn}
Li Yi, Wang Zhao, He Wang, Minhyuk Sung, and Leonidas Guibas.
\newblock Gspn: Generative shape proposal network for 3d instance segmentation
  in point cloud.
\newblock {\em arXiv preprint arXiv:1812.03320}, 2018.

\bibitem{zhao2007robust}
Wei Zhao, Shuming Gao, and Hongwei Lin.
\newblock A robust hole-filling algorithm for triangular mesh.
\newblock {\em The Visual Computer}, 23(12):987--997, 2007.

\end{thebibliography}
}

\newpage
\appendix
\section{Appendix}
In this appendix, we detail our \OURS{} network architecture in Section~\ref{sec:architecture_details}; in Section~\ref{sec:timing}, we provide run-time results of our approach; in Section~\ref{sec:discuss}, we discuss on differences between model-fitting and prediction-based approaches regarding object-level completion in RGB-D scans; in Section~\ref{sec:degree} we show the ablation study that how degrees of completeness in the input data influence the semantic instance complete performance.

\section{Network Architecture}\label{sec:architecture_details}

\begin{table}[h!]
    \centering
     \begin{tabular}{|c|c|} \hline
     small anchors  & big anchors \\ \hline
     (9, 10, 9)    & (47, 20, 23)\\
     (17, 21, 17)  & (23,  20, 47)\\
     (12, 19, 13)  & (16, 18, 30)\\
     (16, 12, 15)  & (17, 38, 17)\\
                   & (30, 18, 16) \\ \hline
    \end{tabular}
    \caption{Anchor sizes used for region proposal on the ScanNet dataset~\cite{dai2017scannet}. Sizes are given in voxel units, with voxel resolution of $\approx 4.69$cm}
    \label{tab:scannet_anchor}
    \vspace{-0.5cm}
\end{table}

\begin{table}[h!]
    \centering
     \begin{tabular}{|c|c|} \hline
     small anchors  & big anchors \\ \hline
     (8, 6, 8)     & (12, 12, 40)\\
     (22, 22, 16)  & (8 , 60, 40)\\
     (12, 12, 20)  & (38, 12, 16)\\
                   & (62, 8 , 40)\\
                   & (46, 8 , 20)\\
                   & (46, 44, 20)\\
                   & (14, 38, 16)\\ \hline
    \end{tabular}
    \caption{Anchor sizes (in voxels) used for SUNCG~\cite{song2017ssc} region proposal. Sizes are given in voxel units, with voxel resolution of $\approx 4.69$cm}
    \label{tab:suncg_anchor}

\end{table}
Table~\ref{tab:backbone} details the layers used in our backbone. 3D-RPN, classification head, and mask completion head are described in Table~\ref{tab:heads}. Additionally, we leverage the residual blocks in our backbone, which is listed in Table~\ref{tab:resblock}.
Note that both the backbone and mask completion head are fully-convolutional. For the classification head, we use several fully-connected layers; however, we leverage 3D RoI-pooling on its input, we can run our method on large 3D scans of varying sizes in a single forward pass.

We additionally list the anchors used for the region proposal for our model trained on our ScanNet-based semantic instance completion benchmark~\cite{avetisyan2019scan2cad,dai2017scannet} and SUNCG~\cite{song2017ssc} datasets in Tables~\ref{tab:scannet_anchor} and \ref{tab:suncg_anchor}, respectively. Anchors for each dataset are determined through $k$-means clustering of ground truth bounding boxes.
The anchor sizes are given in voxels, where our voxel size is $\approx 4.69$cm.

\section{Inference Timing}\label{sec:timing}

In this section, we present the inference timing with and without color projection in Table~\ref{tab:timing} and \ref{tab:ctiming}. Note that our color projection layer currently projects the color signal into 3D space sequentially, and can be further optimized using CUDA, so that it can project the color features back to 3D space in parallel. A scan typically contains several hundreds of images; hence, this optimization could significantly further improve inference time.

\begin{table}[h!]

    \centering
     \begin{tabular}{|c|c|c|c|} \hline
     physical size (m) & 5.8 x 6.4 & 8.3 x 13.9 & 10.9 x 20.1 \\ \hline
     voxel resolution & 124 x 136 & 176 x 296 & 232 x 428 \\ \hline
     forward pass (s)  & 0.15  & 0.37        & 0.72\\ \hline
    \end{tabular}
    \vspace{-0.1cm}
    \caption{Inference time on entire scenes without color signal. Timings are given in seconds, physical sizes are given in meters and spatial sizes are given in voxel units, with voxel resolution of $\approx 4.69$cm}.
    \label{tab:timing}
    \vspace{-0.6cm}
\end{table}

\begin{table}[h!]
\small
    \centering
     \begin{tabular}{|c|c|c|c|} \hline
     physical size (m)   & 4.7 x 7.7 & 7.9 x 9.6 & 10.7 x 16.5 \\ \hline
     voxel resolution   & 100 x 164 & 168 x 204 & 228 x 352 \\ \hline
     image count        & 49      & 107     & 121 \\
     color projection (s)    & 1.43   & 5.16   & 11.78\\ \hline
     forward pass (s)    & 0.19   & 0.34   & 0.64\\ \hline
     total (s)      & 1.62   & 5.50   & 12.42\\ \hline
    \end{tabular}
    \vspace{-0.1cm}
    \caption{Inference timing on entire large scans with RGB input. Timings are given in seconds, physical sizes are given in meters and spatial sizes are given in voxel units, with voxel resolution of $\approx 4.69$cm}.
    \label{tab:ctiming}
    \vspace{-0.5cm}

\end{table}

\section{Model-fitting vs. Prediction-based Methods}
\label{sec:discuss}
In terms of object level completion in the RGB-D scan, We discuss about differences between model-fitting and prediction-based methods. Regarding model-fitting methods, {\em ``Scan2CAD''}~\cite{avetisyan2019scan2cad} and {\em ``End-to-End CAD''}~\cite{avetisyan2019end2end} define a CAD alignment task for which they require a pre-defined set of CAD models for each test scene; i.e., GT objects are given at test time and only alignment needs to be inferred (cf.~Scan2CAD benchmark). Prediction-based method method, e.g. \OURS{},  does not have the GT objects as input in the test time.

Semantic instance completion is fundamentally more flexible as it operates on a per-voxel basis (vs. fixed CAD models); i.e., allowing construction of much more true-to-observation geometry (e.g., necessary in the context of robotics). Since prediction-based methods complete the missing surface based on observed geometry that performs as an anchor, \OURS{} easily overlaps with the ground truth surface, whereas model-fitting methods, which predicts rotation/scale/translation, could have little overlap with the ground truth surface due to slight misalignment (e.g. 10 degrees rotation error).
In addition, we want to highlight that CAD model alignment does not help their detection performance, but instance completion in \OURS{} does.

\begin{table*}[tp!]
    \centering
     \resizebox{0.8\textwidth}{!}{
     \begin{tabular}{|c|c|c|c|c|c|c|c|}\specialrule{1.3pt}{0.0pt}{0.1pt}
      ResBlock  &  Input Layer    &  Type            & Input Size    & Output Size    & Kernel Size & Stride    &  Padding \\ \hline 
      convres0  &  CNN feature    &  Conv3d          & (N,X,Y,Z)     & (N/2,X,Y,Z)    & (1,1,1)     & (1,1,1)   & (0,0,0)  \\
      normres0  &  convres0       &  InstanceNorm3d  & (N/2,X,Y,Z)   & (N/2,X,Y,Z)    & None        & None      & None     \\
      relures0  &  normres0       &  ReLU            & (N/2,X,Y,Z)   & (N/2,X,Y,Z)    & None        & None      & None     \\
      convres1  &  relures0       &  Conv3d          & (N/2,X,Y,Z)   & (N/2,X,Y,Z)    & (3,3,3)     & (1,1,1)   & (1,1,1)  \\
      normres1  &  convres1       &  InstanceNorm3d  & (N/2,X,Y,Z)   & (N/2,X,Y,Z)    & None        & None      & None     \\
      relures1  &  normres1       &  ReLU            & (N/2,X,Y,Z)   & (N/2,X,Y,Z)    & None        & None      & None     \\
      convres2  &  relures1       &  Conv3d          & (N/2,X,Y,Z)   & (N,X,Y,Z)      & (1,1,1)     & (1,1,1)   & (0,0,0)  \\
      normres2  &  convres2       &  InstanceNorm3d  & (N,X,Y,Z)     & (N,X,Y,Z)      & None        & None      & None     \\
      relures2  &  normres2       &  ReLU            & (N,X,Y,Z)     & (N,X,Y,Z)      & None        & None      & None     \\ \hline
    \specialrule{1.3pt}{0.1pt}{0pt}
    \end{tabular}
    }
     \vspace{-0.1cm}
    \caption{Residual block specification in \OURS.}
    \label{tab:resblock}
    \vspace{-0.4cm}
\end{table*}

\section{Performance Study over Degrees of Completeness}
\label{sec:degree}

\begin{figure}[h!]
\begin{center}
   \includegraphics[width=0.95\linewidth]{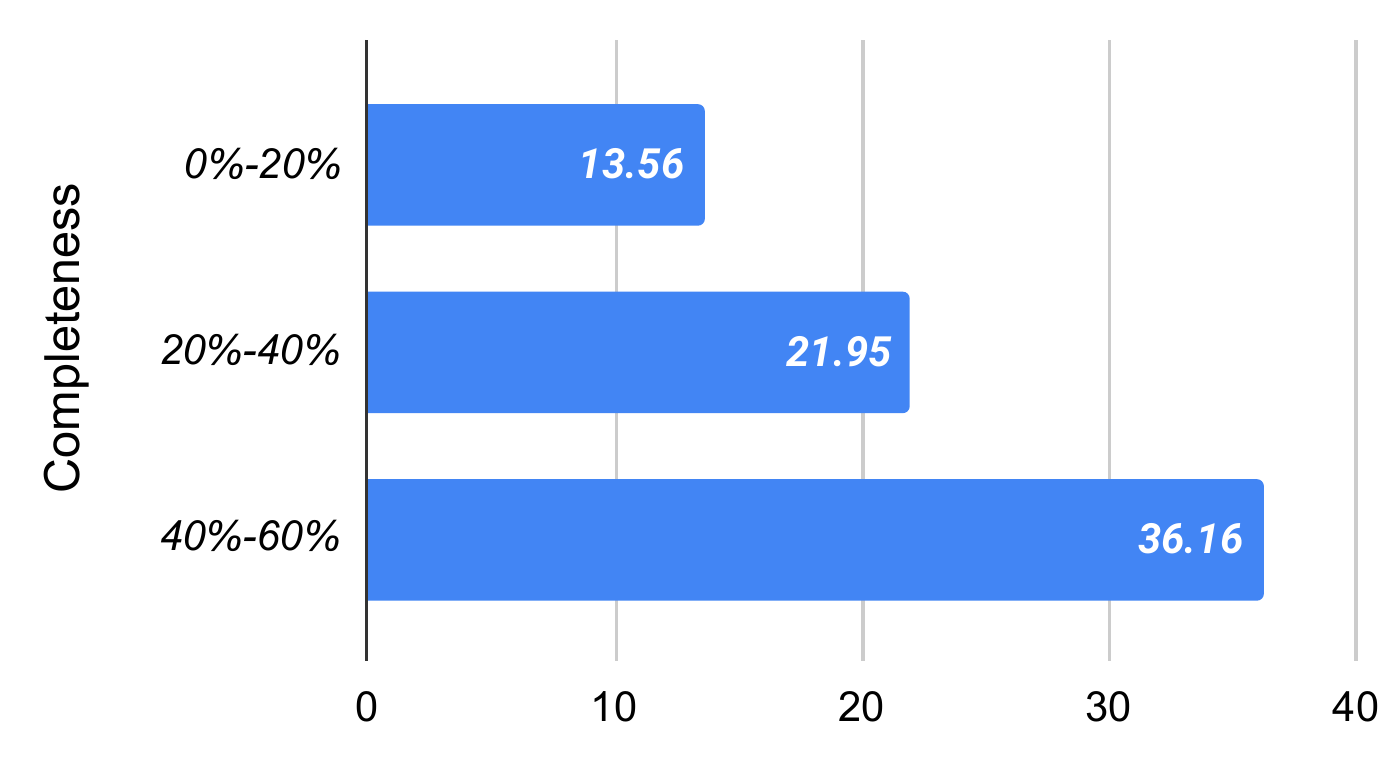}
    \end{center}
    \vspace{-0.1cm}
   \caption{Results of Tab.~1 split by levels of object completeness in mAP@0.5; more complete input is easier.}
    \vspace{-0.5cm}
\label{fig:chart}
\end{figure}

Both our SUNCG and ScanNet settings have a large variety of incompleteness. We show a histogram in Fig.~\ref{fig:chart} of our current results on ScanNet (numbers split from Tab.~1 main paper). From the histogram, We show that with more complete geometry input, semantic instance completion task becomes easier.\\

\begin{table*}[h!]

    \centering
     \resizebox{0.75\textwidth}{!}{
     \begin{tabular}{|c|c|c|c|c|c|c|c|}\specialrule{1.3pt}{0.0pt}{0.1pt}
      BackBone  &  Input Layer    &  Type            & Input Size    & Output Size    & Kernel Size & Stride    &  Padding \\ \hline 
      geometry0 &  TSDF           & Conv3d           & (2,96,48,96)  &(32,48,24,48)   & (2,2,2)     & (2,2,2)   & (0,0,0)  \\
      norm0     &  geometry0      & InstanceNorm3d   & (32,48,24,48) &(32,48,24,48)   & None        & None      &  None    \\
      relu0     &  norm0          & ReLU             & (32,48,24,48) &(32,48,24,48)   & None        & None      &  None    \\ 
      block0    &  relu0          & ResBlock         & (32,48,24,48) &(32,48,24,48)   & None        & None      &  None    \\
              
      color1    &  CNN feature    & Conv3d           & (128,96,48,96)&(32,48,24,48)   & (2,2,2)     & (2,2,2)   & (0,0,0) \\
      norm1     &  color1         & InstanceNorm3d   & (32,48,24,48) &(32,48,24,48)   & None        & None      & None    \\
      relu1     &  norm1          & ReLU             & (32,48,24,48) &(32,48,24,48)   & None        & None      & None    \\ 
      block1    &  relu1          & ResBlock         & (32,48,24,48) &(32,48,24,48)   & None        & None      & None    \\
        
      concat2   & (block0,block1) & Concatenate      & (32,48,24,48) &(64,48,24,48)   & None        & None      & None    \\
      combine2  & concat2         & Conv3d           & (64,48,24,48) &(128,24,12,24)  & (2,2,2)     & (2,2,2)   & (0,0,0) \\
      norm2     & combine2        & InstanceNorm3d   & (128,24,12,24)&(128,24,12,24)  & None        & None      & None    \\
      relu2     & norm2           & ReLU             & (128,24,12,24)&(128,24,12,24)  & None        & None      & None    \\ 
      block2    & relu2           & ResBlock         & (128,24,12,24)&(128,24,12,24)  & None        & None      & None    \\
      
      encoder3  & block2          & Conv3d           & (128,24,12,24)&(128,24,12,24)  & (3,3,3)     & (1,1,1)   & (1,1,1)   \\
      norm3     & combine3        & InstanceNorm3d   & (128,24,12,24)&(128,24,12,24)  & None        & None      & None      \\
      relu3     & norm3           & ReLU             & (128,24,12,24)&(128,24,12,24)  & None        & None      & None      \\ 
      block3    & relu3           & ResBlock         & (128,24,12,24)&(128,24,12,24)  & None        & None      & None      \\
      
      skip4     & (block, block3) & Conv3d           & (128,24,12,24)&(64,48,24,48)   & (2,2,2)     & (2,2,2)   & (0,0,0)   \\
      norm4     & combine4        & InstanceNorm3d   & (64,48,24,48) &(64,48,24,48)   & None        & None      & None      \\
      relu4     & norm4           & ReLU             & (64,48,24,48) &(64,48,24,48)   & None        & None      & None      \\ 
      block4    & relu4           & ResBlock         & (64,48,24,48) &(64,48,24,48)   & None        & None      & None      \\
      
      concat5   & (block2,block4) & Concatenate      & (64,48,24,48) &(128,48,24,48)  & None        & None      & None      \\
      decoder5  & block5          & ConvTranspose3d  & (128,48,24,48)&(32,96,48,96)   & (2,2,2)     & (2,2,2)   & (0,0,0)   \\
      norm5     & combine5        & InstanceNorm3d   & (32,96,48,96) &(32,96,48,96)   & None        & None      & None      \\
      relu5     & norm5           & ReLU             & (32,96,48,96) &(32,96,48,96)   & None        & None      & None      \\ 
      block5    & relu5           & ResBlock         & (32,96,48,96) &(32,96,48,96)   & None        & None      & None      \\ 
      proxy5    & block5          & ConvTranspose3d  & (32,96,48,96) &(1,96,48,96)    & (1, 1, 1)   & (1,1,1)   & (0,0,0)   \\ \hline
    \specialrule{1.3pt}{0.1pt}{0pt}
    \end{tabular}
    }
    \vspace{-0.3cm}
    \caption{Backbone layer specifications in \OURS.}
    \label{tab:backbone}
     \vspace{-0.2cm}
\end{table*}

\begin{table*}[bt!]
    \centering
     \resizebox{0.75\textwidth}{!}{
     \begin{tabular}{|c|c|c|c|c|c|c|c|}\specialrule{1.3pt}{0.0pt}{0.1pt}
      RPN       &  Input Layer    &  Type            & Input Size    & Output Size    & Kernel Size & Stride    &  Padding  \\ \hline 
      rpn6      & block2          & Conv3d           & (128,24,12,24)&(256,24,12,24)  & (3,3,3)     & (1,1,1)   & (1,1,1)   \\
      norm6     & rpn6            & InstanceNorm3d   & (256,24,12,24)&(256,24,12,24)  & None        & None      & None      \\
      relu6     & norm6           & ReLU             & (256,24,12,24)&(256,24,12,24)  & None        & None      & None      \\ 
      
      rpncls7a  & relu6           & Conv3d           & (256,24,12,24)&(8,24,12,24)    & (1,1,1)     & (1,1,1)   & (0,0,0)   \\
      norm7a    & rpncls7a        & InstanceNorm3d   & (8,24,12,24)  &(8,24,12,24)    & None        & None      & None      \\
      rpnbbox7b & relu6           & Conv3d           & (24,24,12,24) &(24,24,12,24)   & (1,1,1)     & (1,1,1)   & (0,0,0)   \\
      norm7b    & rpnbbox7b       & InstanceNorm3d   & (24,24,12,24) &(24,24,12,24)   & None        & None      & None      \\
      
      rpn8      & block3          & Conv3d           & (128,24,12,24)&(256,24,12,24)  & (3,3,3)     & (1,1,1)   & (1,1,1)   \\
      norm8     & rpn8            & InstanceNorm3d   & (256,24,12,24)&(256,24,12,24)  & None        & None      & None      \\
      relu8     & norm8           & ReLU             & (256,24,12,24)&(256,24,12,24)  & None        & None      & None      \\ 
      
      rpncls9a  & relu8           & Conv3d           & (256,24,12,24)&(8,24,12,24)    & (1,1,1)     & (1,1,1)   & (0,0,0)   \\
      norm9a    & rpncls9a        & InstanceNorm3d   & (10,24,12,24) &(10,24,12,24)   & None        & None      & None      \\
      rpnbbox9b & relu8           & Conv3d           & (30,24,12,24) &(30,24,12,24)   & (1,1,1)     & (1,1,1)   & (0,0,0)   \\
      norm9b    & rpnbbox9b       & InstanceNorm3d   & (30,24,12,24) &(30,24,12,24)   & None        & None      & None      \\ \hline
      
      Class Head&  Input Layer    &  Type            & Input Size    & Output Size    & Kernel Size & Stride    &  Padding  \\ \hline 
      roipool10 &block2/block3    & RoI Pooling      & (64,arbitrary)& (64, 4, 4, 4)  & None        & None      & None      \\
      flat10    &roipool10        & Flat             & (64,4,4,4)    & (4096)         & None        & None      & None      \\
      cls10a    & flat10          & Linear           & (4096)        & (256)          & None        & None      & None      \\
      relu10a   & cls10a          & ReLU             & (256)         & (256)          & None        & None      & None      \\ 
      cls10b    & relu10a         & Linear           & (256)         & (128)          & None        & None      & None      \\ 
      relu10b   & cls10b          & ReLU             & (128)         & (128)          & None        & None      & None      \\ 
      cls10c    & relu10b         & Linear           & (128)         & (128)          & None        & None      & None      \\ 
      relu10c   & cls10c          & ReLU             & (128)         & (128)          & None        & None      & None      \\ 
      clscls10  & relu10c         & Linear           & (128)         & (8)            & None        & None      & None      \\ 
      clsbbox10 & relu10c         & Linear           & (128)         & (48)           & None        & None      & None      \\ \hline
      Mask Head&  Input Layer     &  Type            & Input Size    & Output Size    & Kernel Size & Stride    &  Padding  \\ \hline 
      mask11     & block2/block3  & Conv3d           & (N,arbitrary) &(N,arbitrary)   & (9,9,9)     & (1,1,1)   & (4,4,4)   \\
      norm11     & mask11         & InstanceNorm3d   & (N,arbitrary) &(N,arbitrary)   & None        & None      & None      \\
      relu11     & norm11         & ReLU             & (N,arbitrary) &(64,arbitrary)  & None        & None      & None      \\ 
      
      mask12     & relu11         & Conv3d           & (N,arbitrary) &(N,arbitrary)   & (7,7,7)     & (1,1,1)   & (3,3,3)   \\
      norm12     & mask12         & InstanceNorm3d   & (N,arbitrary) &(N,arbitrary)   & None        & None      & None      \\
      relu12     & norm12         & ReLU             & (N,arbitrary) &(64,arbitrary)  & None        & None      & None      \\ 
      
      mask13     & relu12         & Conv3d           & (N,arbitrary) &(N,arbitrary)   & (5,5,5)     & (1,1,1)   & (2,2,2)   \\
      norm13     & mask13         & InstanceNorm3d   & (N,arbitrary) &(N,arbitrary)   & None        & None      & None      \\
      relu13     & norm13         & ReLU             & (N,arbitrary) &(64,arbitrary)  & None        & None      & None      \\ 
      
      mask14     & relu13         & Conv3d           & (N,arbitrary) &(N,arbitrary)   & (3,3,3)     & (1,1,1)   & (1,1,1)   \\
      norm14     & mask14         & InstanceNorm3d   & (N,arbitrary) &(N,arbitrary)   & None        & None      & None      \\
      relu14     & norm14         & ReLU             & (N,arbitrary) &(64,arbitrary)  & None        & None      & None      \\ 
      
      mask15     & relu14         & Conv3d           & (N,arbitrary) &(N,arbitrary)   & (1,1,1)     & (1,1,1)   & (0,0,0)   \\
      
    \specialrule{1.3pt}{0.1pt}{0pt}
    \end{tabular}
    }
    \vspace{-0.3cm}
    \caption{Head layer specifications of RPN, Classification and Mask Completion in \OURS.}
    \label{tab:heads}
    
\end{table*}

\end{document}